\documentclass{article}

\usepackage{microtype}
\usepackage{graphicx}
\usepackage{subcaption}
\usepackage{booktabs} 

\usepackage{hyperref}
\usepackage{subcaption}
\usepackage{graphicx}

\usepackage[preprint]{icml2026}

\usepackage{amsmath}
\usepackage{amssymb}
\usepackage{mathtools}
\usepackage{amsthm}

\usepackage{enumitem}

\newcommand{\eg}{\textit{e.g.}}

\newcommand{\ie}{\textit{i.e.}}

\usepackage{pifont}

\usepackage[capitalize,noabbrev]{cleveref}

\theoremstyle{plain}

\theoremstyle{definition}

\theoremstyle{remark}

\usepackage[most,skins,theorems]{tcolorbox}
\tcbset{
  aibox/.style={
    width=\linewidth,
    top=2pt,
    bottom=2pt,
    colback=blue!6!white,
    colframe=black,
    colbacktitle=black,
    enhanced,
    center,
    attach boxed title to top left={yshift=-0.1in,xshift=0.15in},
    boxed title style={boxrule=0pt,colframe=white,},
  }
}
\newtcolorbox{AIbox}[2][]{aibox,title=#2,#1}

\usepackage[textsize=tiny]{todonotes}

\icmltitlerunning{Submission and Formatting Instructions for ICML 2026}

\begin{document}

\twocolumn[
  \icmltitle{Position: The Real Barrier to LLM Agent Usability is Agentic ROI}



  \icmlsetsymbol{corr}{*}


\begin{icmlauthorlist}
    \icmlauthor{Weiwen Liu}{sjtu}
    \icmlauthor{Jiarui Qin}{}
    \icmlauthor{Xu Huang}{ustc}
    \icmlauthor{Xingshan Zeng}{corr}
    \icmlauthor{Yunjia Xi}{sjtu}
    \icmlauthor{Jianghao Lin}{sjtu}
    \icmlauthor{Chuhan Wu}{corr}
    \icmlauthor{Yasheng Wang}{corr}
    \icmlauthor{Lifeng Shang}{}
    \icmlauthor{Ruiming Tang}{}
    \icmlauthor{Defu Lian}{ustc}
    \icmlauthor{Yong Yu}{sjtu}
    \icmlauthor{Weinan Zhang}{sjtu,corr}
  \end{icmlauthorlist}

  \icmlaffiliation{sjtu}{Shanghai Jiao Tong University}
  \icmlaffiliation{ustc}{University of Science and Technology of China}

  \icmlcorrespondingauthor{Weiwen Liu}{wwliu@sjtu.edu.cn}

  \icmlkeywords{Machine Learning, ICML}

  \vskip 0.3in
]



\printAffiliationsAndNotice{*Corresponding Author.}  

\begin{abstract}
Large Language Model (LLM) agents represent a promising shift in human-AI interaction, moving beyond passive prompt-response systems to autonomous agents capable of reasoning, planning, and goal-directed action. 
While LLM agents are technically capable of performing a broad range of tasks, not all of these capabilities translate into meaningful usability.
This position paper argues that \textbf{the central question for LLM agent usability is no longer whether a task \textit{can} be automated, but whether it delivers sufficient Agentic Return on Investment (Agentic ROI).}
Agentic ROI reframes evaluation from raw performance to a holistic, utility-driven perspective, guiding \textit{when}, \textit{where}, and \textit{for whom} LLM agents should be deployed. 
Despite widespread application in high-ROI tasks like coding and scientific research, we identify a critical usability gap in mass-market, everyday applications.
To address this, we propose a zigzag developmental trajectory: first scaling up to improve information gain and time savings, then scaling down to reduce cost. 
We present a strategic roadmap across these phases to make LLM agents truly usable, accessible, and scalable in real-world applications.
\end{abstract}

\section{Introduction}
The year 2025 marked an inflection point for Large Language Model (LLM) agents. Foundation models like GPT-5~\cite{openai2026gpt5}, Gemini-3~\cite{google2025gemini}, Qwen-3~\cite{yang2025qwen3}, and DeepSeek-V3.2~\cite{liu2025deepseek} established a robust backbone for agentic behavior, while protocols like MCP~\cite{anthropic2024mcp} and A2A~\cite{google2025a2a} standardized agent coordination. Coupled with agent workflow orchestration tools like n8n~\cite{n8n2025}, these advances transitioned AI from static prompt-response chatbots to autonomous, goal-directed systems~\cite{yang2025survey}.

Consequently, interest in deploying agents has surged across domains ranging from software development~\cite{novikov2025alphaevolve}, education~\cite{chu2025llm}, healthcare~\cite{yang2025medaide}, to scientific research~\cite{huang2025deep}. Notably, over 90\% of Salesforce engineers now use Cursor, achieving double-digit gains in velocity and quality~\cite{cursor2026salesforce}. However, mass-market deployment remains sparse compared to established AI systems. While algorithmic recommendation engines drive over 700 million MAUs on platforms like Douyin~\cite{business2025tiktok}, advanced agent services like OpenAI Plus reach only 10 million users~\cite{demandsage2025chatgpt}, indicating a substantial untapped market.

In practice, current LLM agents are predominantly deployed in specialized, high-effort domains like coding or research~\cite{wermelinger2023using,openai2025introducing}, where the typical users are already domain experts and the burden of iterative prompt engineering is justified by the high value of the outcomes. This leaves LLM agents largely inaccessible to the general public. We believe that this discrepancy arises because purely optimizing agent performance overlooks the broader socio-technical ecosystem, creating a critical usability gap.

To bridge this gap, \textbf{we argue that the usability of LLM agents depends not solely on raw capability, but on maximizing Agentic Return on Investment (Agentic ROI).}  Agentic ROI quantifies the utility an agent provides to the user relative to the costs incurred during real-world use. 
Crucially, Agentic ROI is context-dependent and personalized.
By combining both human cognitive load and agent performance into a single equation, Agentic ROI provides the first holistic evaluation of LLM agent usability.
It serves as a strategic guide for identifying \textit{when} (at what maturity level), \textit{where} (in which specific workflows), and \textit{for whom} (which user expertise levels) to deploy LLM agents.
Therefore, we define the optimization of Agentic ROI as the primary objective for the next generation of agentic systems.

In the sections that follow, we first formalize the theoretical concept of Agentic ROI (Section~\ref{sect:agentic_roi}). Next, we demonstrate how it can be practically operationalized, and analyze its implications for the current usability gap (Section~\ref{sect:survey}). Building on this, we propose a development roadmap for optimizing Agentic ROI (Section~\ref{sect:zigzag}). After discussing alternative perspectives (Section~\ref{sect:alternative}), we conclude in Section~\ref{sect:conclusion}. Related literature and broader impacts appear in the Appendix.

\section{Agentic ROI}\label{sect:agentic_roi}
To systematically evaluate the usability of deploying LLM agents in real-world applications, we introduce the metric of Agentic ROI. 
This metric serves as a composite index of economic leverage, quantifying the utility generated by an agent relative to the cost.
We define Agentic ROI as:
\begin{AIbox}{\textbf{Agentic ROI}}
\begin{align}
\label{eq:roi}
    &\text{Agentic ROI} \\=& \frac{\text{Information Gain} \times \text{Time Savings}}{\text{Cost}}\notag\\
    =&\frac{\text{max}(Q_{Agent}-Q_0,0)\times\text{max}(T_0-T_{Agent},0)}{{\text{Cost}}}\,.\notag
\end{align}
\end{AIbox}
Specifically, Agentic ROI computes the compound value derived from the agent's ability to improve output quality and reduce task completion time compared to a human baseline, normalized by the total cost of applying the agent, where $\textit{Agentic ROI}\in[0,+\infty)$.
A higher ROI indicates better usability of LLM agents for the corresponding domain.
The formulation is composed of three components:
\begin{itemize}[leftmargin=*]
    \item \textbf{Information Gain} captures the improvement in output quality delivered by the agent over a human baseline. Here, $Q_{Agent}$ refers to the quality of the agent (\ie, accuracy, usefulness, and completeness), while $Q_0$ represents the baseline human quality, measuring the minimum acceptable standard for human performance. The function $\text{max}(\cdot,0)$ acts as a filter: if an agent performs worse than the human baseline ($Q_{Agent} < Q_0$), the information gain is zero, resulting in zero Agentic ROI regardless of time savings. In practice, information gain can be assessed through user ratings or proxy metrics on standardized scales (\eg, 0–1, 1–5).
    \item \textbf{Time Savings} quantifies the reduction in human hours when using the agent, where $T_0$ represents the time a human would require to complete the task independently, and $T_{Agent}$ is the total time the human spends interacting with the agent to accomplish the task. This includes time spent on prompt formulation, intent clarification, iterative refinements, and result verification. The difference $T_0-T_{Agent}$ thus captures the net time efficiency gained through agent-assisted execution.
    \item \textbf{Cost} refers to the monetary cost incurred (\eg, token consumption, API fees, or compute resources), with $\textit{Cost}>0$.
\end{itemize}

Overall, Agentic ROI offers a principled framework for evaluating the value of deploying LLM agents in diverse real-world settings. It can be computed at varying levels of granularity---such as across different user populations, domains, or system configurations---enabling detailed analysis of agent performance. In practical deployment scenarios, Agentic ROI supports informed decision-making about when, where, and for whom LLM agents should be applied to maximize impact and cost-efficiency.

One of the key features of the Agentic ROI framework is its \textit{personalizability}. Since both the human baseline quality $Q_0$ and task completion time $T_0$ are user-dependent, Agentic ROI naturally varies across individuals. For instance, users with lower baseline performance, such as those with disabilities, limited domain expertise, or lower digital literacy, may experience significant benefits from even modest improvements in task quality or efficiency. In such cases, the resulting Agentic ROI can be disproportionately high. 

\section{Empirical Observations of Agentic ROI}
In this section, we provide a demonstration of practically computing and analyzing Agentic ROI. 
\label{sect:survey}
\subsection{Agentic ROI Quantification}
To quantify Agentic ROI, we collected a total of 34 valid survey responses from individuals who regularly interact with LLM agents. The sample comprised 14 AI practitioners and 20 end-users, ensuring a balanced perspective across both technical and non-technical audiences. All responses were gathered via an online questionnaire, with questions randomized at the domain level to mitigate potential order effects.

Our analysis focuses on five primary application domains of LLM agents: 
(1)~\textit{Coding}: Agents that facilitate software development via code generation, intelligent autocompletion, and automated debugging assistance~\cite{novikov2025alphaevolve,robeyns2025self}. 
(2)~\textit{Research}: Agents designed to accelerate information discovery, automate literature reviews, and support scientific reasoning~\cite{huang2025deep,open_deep_research,zheng2025deepresearcher}. 
(3)~\textit{Office Work}: Agents integrated into enterprise workflows to execute professional productivity tasks~\cite{zheng2025pptagent,microsoft365copilot2026}. 
(4)~\textit{E-commerce}: Agents that optimize online retail through automated customer support, recommendation engines, and virtual shopping assistance~\cite{xiao2025mmagentrec}. 
(5)~\textit{Personal Assistance}: Agents that help individuals manage daily tasks (\eg, health management, travel planning)~\cite{wang2025mobile, zhang2025appagent}.
For each application domain, participants evaluated their general experience with agents by responding to the following items:

\textit{\textbf{(Q1) (Information Gain)} Compared to a purely human workflow, to what extent do agents in this domain improve output quality? 10-point scale (1 = “Much worse,” 10 = “Much better”).}

\textit{\textbf{(Q2) (Time Savings)} On average, how many minutes per task do you save by using agents in this domain? (Time savings = time to complete the task manually – total time the human spent using the agent, including prompting, review, and correction. Numeric entry in minutes).}

\textit{\textbf{(Q3) (Usability)} How would you rate the overall usability of agents in this domain? 10-point scale (1 = “Very difficult to use,” 10 = “Very easy to use”). Please provide an explanation for your rating.}

\begin{figure}[t!]
    \centering
    \begin{subfigure}[b]{0.8\linewidth}
        \centering
        \includegraphics[width=\linewidth]{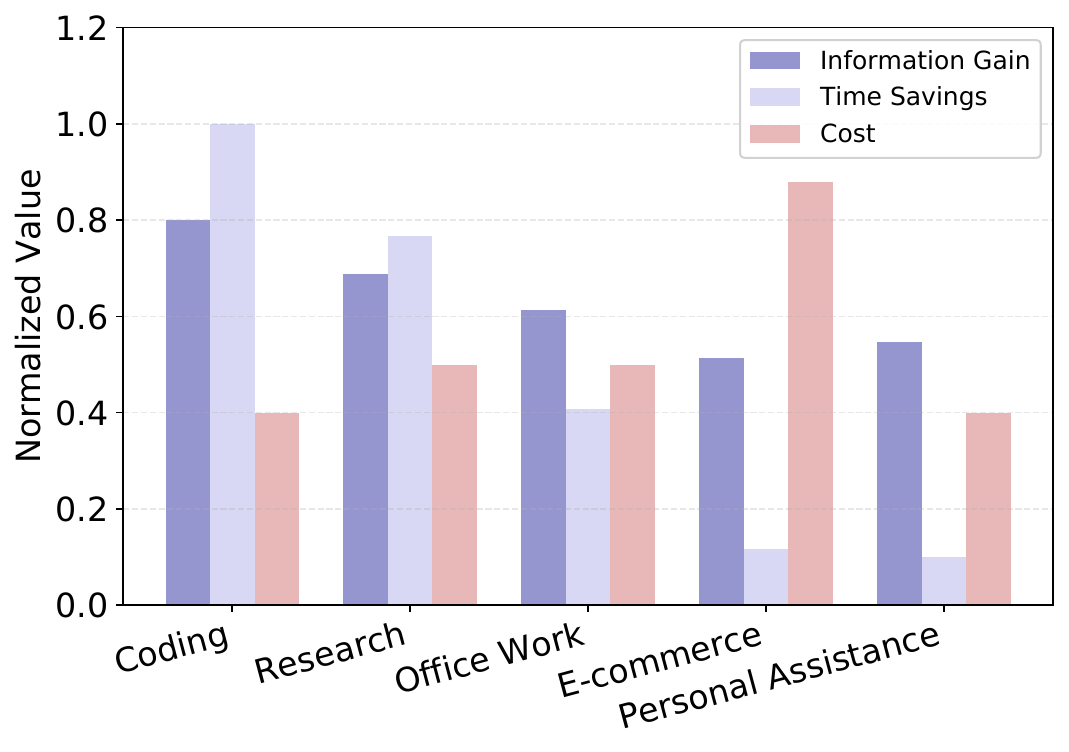}
        \caption{Results of information gain, time savings, and cost.}
        \label{fig:bar}
    \end{subfigure}
    \hfill 
    \begin{subfigure}[b]{0.8\linewidth}
        \centering
        \includegraphics[width=\linewidth]{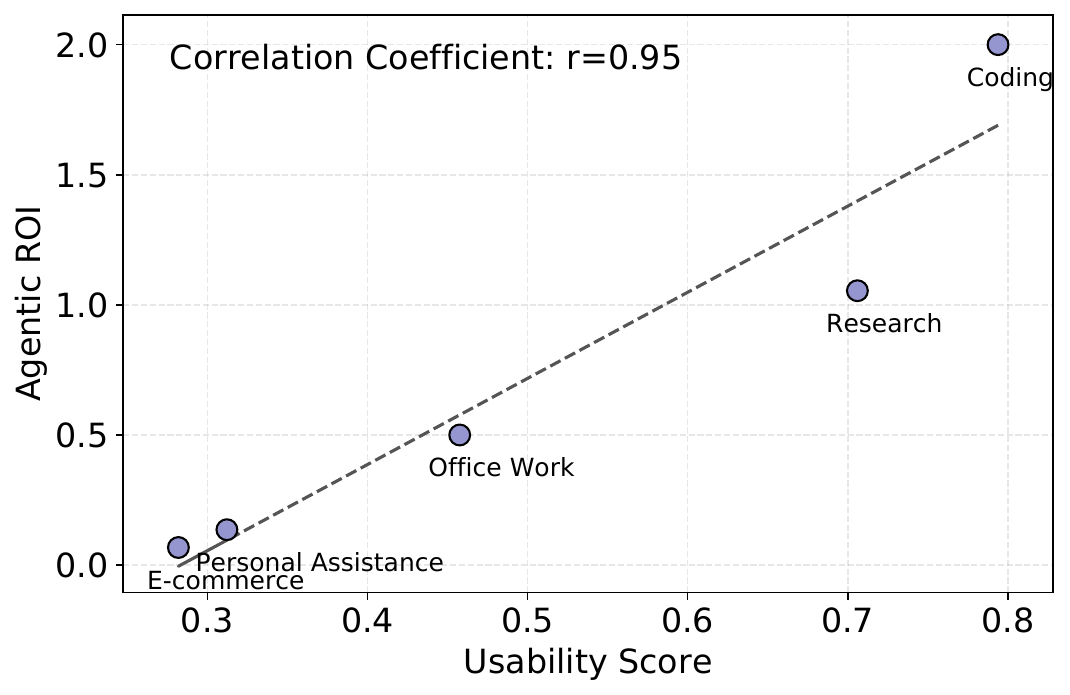}
        \caption{Correlation between Agentic ROI and usability.}
        \label{fig:correlation}
    \end{subfigure}
    \caption{Survey result analysis.}
    \label{fig:survey_analysis}
\end{figure}

To compute Agentic ROI, the raw scores for \textit{Information Gain} and \textit{Time Savings} were averaged across all 34 participants for each domain and subsequently normalized to the range $[0, 1]$. For the \textit{Cost} metric, we estimated the per-task expenditure by dividing the monthly subscription fee of premium agent services by the maximum allowable monthly tasks; these values were similarly normalized to $[0, 1]$. The resulting comparative analysis is presented in Figure~\ref{fig:bar}.

We further examine the relationship between the domain-level average usability reported by participants and the computed Agentic ROI (as defined in Eq~\eqref{eq:roi}). A strong positive linear correlation ($r = 0.95$) is observed in Figure~\ref{fig:correlation}, empirically validating our hypothesis that agent usability is intrinsically linked to realized Agentic ROI.

\subsection{Domain-Specific Trends}

\textbf{High-ROI Domains:} As shown in Figure~\ref{fig:survey_analysis}, domains such as software development and scientific research consistently yield the highest Agentic ROI, with notable gains along both key dimensions: \textit{Information Gain} and \textit{Time Savings}. These domains are characterized by a high baseline human effort $T_0$, where tasks often involve hours of reading, writing, debugging, or analytical reasoning. In such contexts, even partial task automation or augmentation can lead to substantial value.
Moreover, agent performance in these domains is becoming increasingly competitive. Large-scale benchmarks and real-world deployments show that LLM agents are beginning to match, and in some cases exceed, the productivity of skilled human developers and researchers~\cite{google2025gemini}. This contributes to high agent quality $Q_{Agent}$, further amplifying ROI. 

\textbf{Low-ROI Domains:}
In contrast, domains such as office work, e-commerce, and personal assistance currently demonstrate relatively low Agentic ROI. These tasks typically involve short, well-structured interactions (\eg, clicking or swiping), where the baseline human time $T_0$ is already minimal due to years of UX optimization~\cite{hamidli2023introduction} and the user's familiarity with the context. 
In these cases, the time required to instruct or clarify the task to the agent $T_{Agent}$, including context setup, disambiguation, or corrections, can paradoxically exceed the time needed for manual execution. This results in negligible ROI, despite the agent’s theoretical capability.
For example, in office workflows, describing a task in natural language (\eg, ``Schedule a 30-minute sync with the team next week avoiding Thursdays'') often requires a back-and-forth exchange to resolve ambiguities. In contrast, the same task can be completed in seconds using a modern calendar UI. Similarly, in e-commerce, users can browse, filter, and purchase with a few taps, leaving little room for agents to improve on this already efficient process.

This reveals a deeper usability barrier in low-ROI domains: the problem is not a lack of capability, but a lack of marginal gain per cost. Simply improving model accuracy will not meaningfully improve usability in these contexts unless the underlying development strategies and interaction paradigms are reimagined.

\begin{figure*}[t]
\vspace{-5pt}
    \centering
    \includegraphics[width=0.85\linewidth]{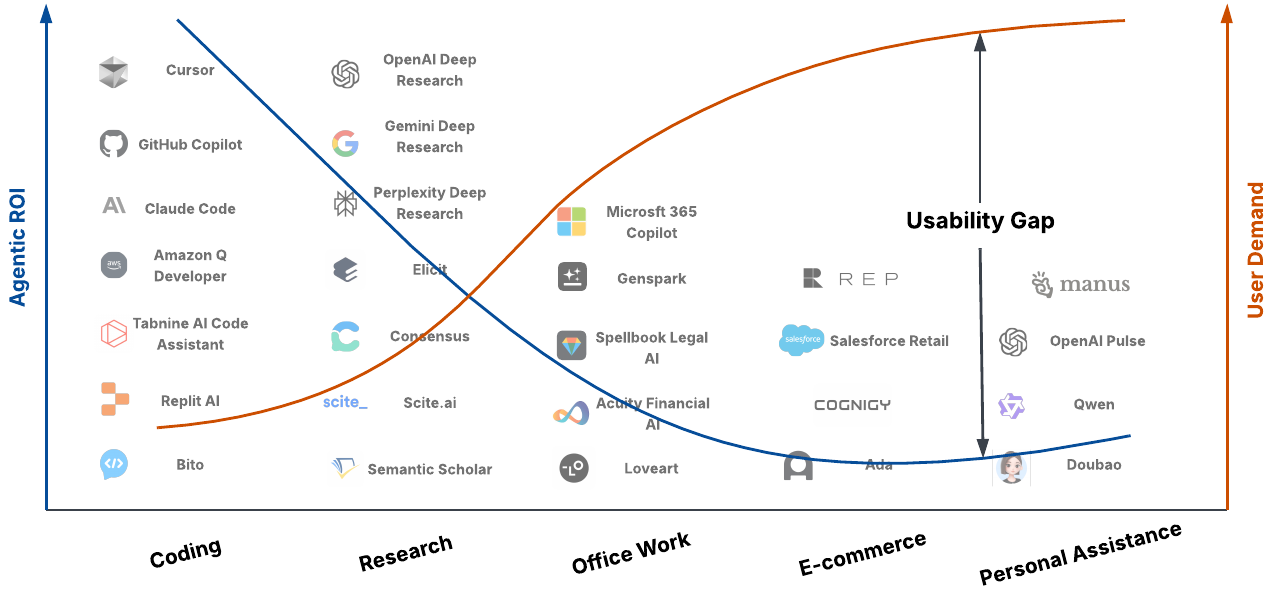}
    \caption{The landscape of LLM agent deployment: User Demand vs. Agentic ROI. User demand is proxied by the monthly active users (MAU) of conventional applications in each domain. Agentic ROI represents conceptual relative trends. Listed agent products are illustrative and non-exhaustive.}
    \label{fig:agentic_roi}
    \vspace{-5pt}
\end{figure*}

\subsection{The Usability Gap}\label{sect_usability_trend}
To assess the current deployment landscape of LLM agents, we analyze the relationship between Agentic ROI and market demand across five representative application domains. As shown in Figure~\ref{fig:agentic_roi}, market demand is estimated using publicly available data on Monthly Active Users (MAU) for conventional applications, serving as a proxy for user engagement and the potential scale of agent adoption~\cite{business2025tiktok}.

From this analysis, we observe an inverse relationship: \textbf{domains with the highest user demand currently exhibit low Agentic ROI, exposing a critical usability gap.}
For instance, domains such as e-commerce, personal assistance, and office work represent some of the largest user bases globally, with hundreds of millions to billions of users interacting with digital tools daily~\cite{wang2022recommendation}. Yet, these domains show limited agent adoption mainly because for a non-technical user, a single swipe or click is often faster and less cognitively demanding than drafting a complex prompt.

This disparity suggests that the bottleneck to mass-market deployment is no longer raw reasoning capability, but the real-world cost-benefit tradeoff as experienced by users. 
Notably, this gap persists even as models achieve high performance on agentic benchmarks such as AndroidWorld~\cite{rawles2024androidworld} and GAIA~\cite{mialon2023gaia}.

To better understand this disconnect, we analyze participants’ explanations for their agent usability ratings. Several key patterns emerge:
\begin{itemize}[leftmargin=*]
    \item \textbf{Agent latency is not a primary barrier to usability.} Users reported a high tolerance for response delays. As long as the agent produces helpful outputs, moderate latency is generally considered acceptable. Given that users often multitask during these intervals, this finding suggests the viability of further scaling test-time computation to improve output quality without significantly impacting perceived usability.
    \item \textbf{Composite, contextually demanding tasks amplify Agentic ROI.} Users consistently rated agents as more valuable in domains involving multi-step workflows where the baseline human time $T_0$ is high. These composite tasks often require sustained coordination, creating greater opportunities for agentic assistance. Importantly, $T_0$ is not static; it varies significantly with context. For example, ordering a coffee typically has a low $T_0$ in everyday settings, but that same task becomes effectively intractable (\ie, unbounded $T_0$) when the user is driving or otherwise unable to interact manually. In such scenarios, the agent’s ability to act autonomously transforms even simple tasks into high-leverage opportunities.
    \item \textbf{Prompting remains a key usability challenge.} Participants noted that formulating effective prompts acts as a significant barrier to entry. Crafting instructions that yield reliable, high-quality outputs often requires trial and error, domain knowledge, or familiarity with the agent’s limitations. This prompting overhead introduces cognitive load that can offset potential time savings, highlighting the need for more intuitive interaction paradigms.
\end{itemize}


\section{Zigzag Development Trend in Optimizing Agentic ROI}\label{sect:zigzag}
\begin{figure}[t]
\vspace{-5pt}
    \centering
    \includegraphics[width=0.8\linewidth]{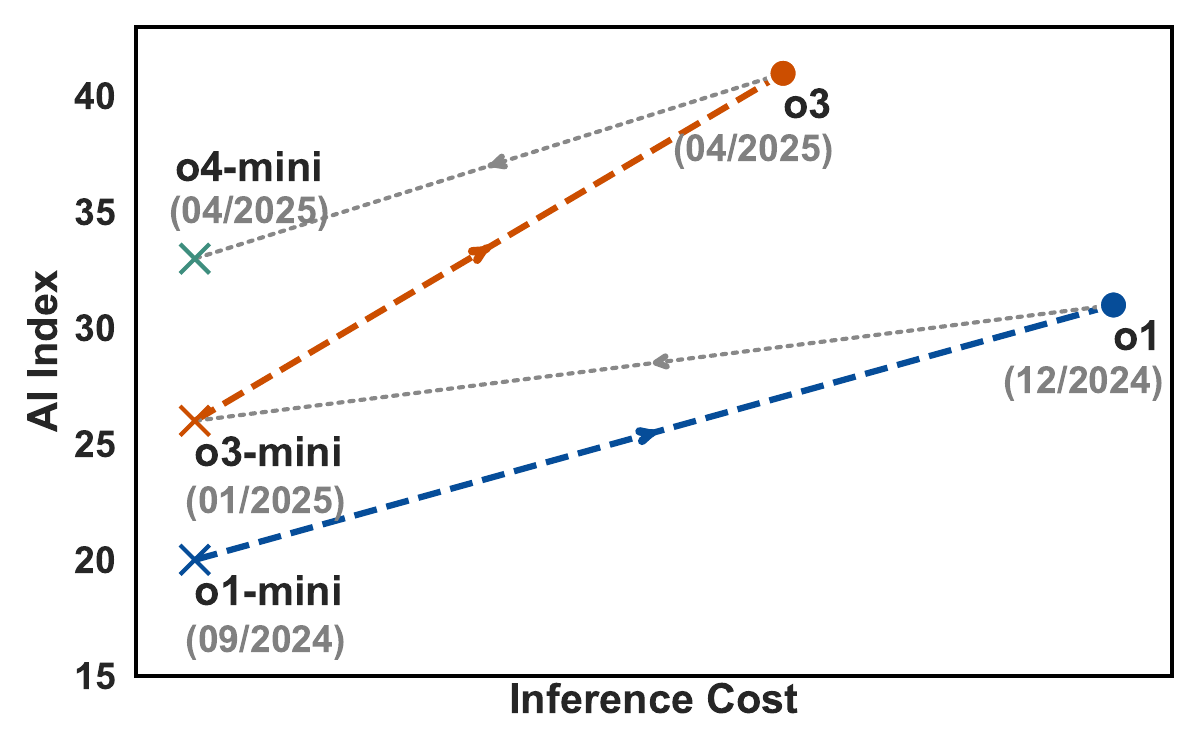}
    \caption{Zigzag performance trends of OpenAI models. The AI Index measures holistic capability by aggregating ten benchmarks across math, science, coding, and reasoning~\cite{maslej2025artificial}.}
    \label{fig:s_vs_p}
    \vspace{-20pt}
\end{figure}
Building on these findings, we argue that \textbf{the development and deployment of LLM agents should explicitly optimize for Agentic ROI}.
This leads us to the next critical question: \textit{how can Agentic ROI be effectively optimized?}

We posit that the trajectory of LLM agent development is unlikely to follow a linear path. Instead, it may exhibit a \textit{zigzag} pattern characterized by alternating phases of scaling up and scaling down. This insight is inspired by performance trends observed in the OpenAI model series (see Figure~\ref{fig:s_vs_p}). Specifically, we identify a recurring cycle: new model families typically begin by \textit{scaling up} to push the frontier of capabilities (\eg, the transition from \texttt{o1-mini} to \texttt{o1}). Subsequent iterations then tend to \textit{scale down}, introducing smaller, more efficient variants (\eg, \texttt{o3-mini}, \texttt{o4-mini}) that retain comparable performance while significantly reducing inference costs (\eg, \texttt{o3-mini} vs. \texttt{o1}, \texttt{o4-mini} vs. \texttt{o3}).

In the context of LLM agents, we observe a similar optimization process in which each generation makes tradeoffs along different axes of the ROI surface: information gain, time savings, and cost---\textbf{first scaling up to improve information gain and time savings, then scaling down to reduce cost.} 
This zigzag development trend is not unique to LLM agents. Similar patterns have been observed in other areas of technological innovation, \eg, computer processors and smartphones~\cite{lotfi2012scale}.

At present, the development of LLM agents remains in the ``scaling up'' phase. 
Large-scale models are being deployed to push the upper bounds of generalization, reasoning, and tool usage~\cite{google2025gemini,openai2026gpt5}.
The high information quality and time savings of LLM agents are achieved at the cost of increased computation and infrastructure demands.
Looking ahead, we anticipate a ``scaling down'' phase in which advances in efficiency, specialization, and model architecture will lower the cost of deployment.

Recognizing this zigzag developmental trajectory allows us to align design principles and expectations with the current phase of LLM agent evolution. Moreover, it underscores the importance of Agentic ROI as a guiding metric---one that captures the dynamic tradeoffs between short-term costs and long-term scalability.

\begin{figure*}
\vspace{-5pt}
    \centering
    \includegraphics[width=0.85\linewidth]{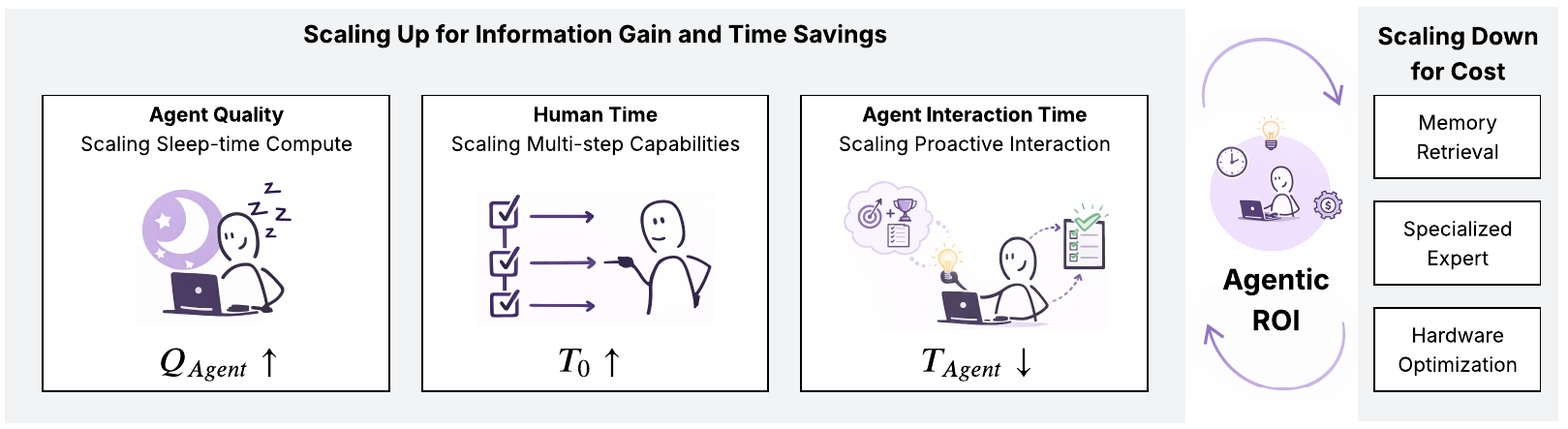}
    \caption{The optimization roadmap for Agentic ROI. The cycle alternates between the scaling up and scaling down phases.}
    \label{fig:optimize}
    \vspace{-15pt}
\end{figure*}

In the following section, we propose key design principles for navigating this zigzag path and outline potential development strategies aimed at maximizing Agentic ROI. The optimization roadmap is shown in Figure~\ref{fig:optimize}.

\subsection{Scaling Up for Information Gain and Time Savings}
The current phase of LLM agent development prioritizes scaling up for information gain and time savings.
Building on our earlier findings, we identify key strategies for scaling up that directly contribute to optimizing Agentic ROI. These strategies span three critical dimensions: sleep-time compute, multi-step capabilities, and proactive interaction. Each targets a core component of the ROI equation---improving agent quality $Q_{Agent}$, capturing high-value human tasks (high $T_0$), and minimizing interaction time $T_{Agent}$.

\subsubsection{Agent Quality: Scaling Sleep-time Compute}
Inspired by the observation in Section~\ref{sect_usability_trend} that users often multi-task during agent execution, we propose that LLM agents can leverage this parallel workflow to generate higher-quality results. Specifically, scaling sleep-time compute~\cite{lin2025sleep} offers a compelling alternative to traditional test-time scaling. By shifting the computational burden from the moment of interaction to the agent's idle periods, we can improve agent quality $Q_{Agent}$ without increasing response latency.

Sleep-time compute effectively decouples task completion into two phases: \textit{offline user understanding} and \textit{online serving}. Currently, a user's multi-modal interaction history contains rich, personalized preferences that remain underutilized~\cite{resnick1997recommender}. Sleep-time compute provides a new human-agent collaboration paradigm where the agent proactively evolves during downtime rather than remaining static between prompts. By continuously refining their internal user models offline, LLM agents can anticipate needs and tailor responses to deliver relevant, personalized outputs in real time.

While OpenAI Pulse provides a preliminary demo of sleep-time compute by synthesizing memory and chat history to prepare a morning briefing~\cite{openai2024pulse}, it largely limits the scope to summarization and retrieval. However, we argue that the true potential of agentic compute lies in \textit{scaling} the depth of sleep-time reasoning. Users' underlying intents are reflected in their history not merely as explicit facts, but as latent clues requiring deep computational effort to uncover.
To fully realize the benefits of sleep-time compute, we identify three key scaling dimensions:
\begin{itemize}[leftmargin=*]
\item \textbf{Scaling user preference reasoning:} Most current systems rely on surface-level cues, such as explicit queries, to infer user needs~\cite{li2024personal}. However, users’ preferences, values, and goals are often implicit, buried within years of fragmented digital traces~\cite{chai2025longer}. In many cases, users themselves may struggle to articulate their underlying intents~\cite{mohamed2019recommender}.
Sleep-time scaling enables agents to perform deep, multi-hop reasoning over historical data without latency constraints. By scaling both the depth (long-range temporal reasoning) and breadth (cross-modal correlation) of preference reasoning, agents shift from passive responses to a deeper, contextualized understanding of the user.

\item \textbf{Scaling iterative simulation:} Idle periods offer a critical window for agents to engage in extensive self-play and counterfactual simulation~\cite{lin2025sleep}. By generating divergent future trajectories and validating them against live APIs or external data sources~\cite{fang2025towards,froger2025scaling}, agents can pre-compute optimal solutions and update their operational knowledge. Consequently, this ``mental sandbox'' enables the system to produce more robust, verified strategies.

\item \textbf{Scaling agent swarms:} Sleep-time compute provides the ideal environment to deploy agent swarms~\cite{kimi2026k25}, collaborative networks of specialized sub-agents, without the bottleneck of real-time coordination latency. During idle periods, a primary agent can decompose complex, latent user goals into sub-problems and distribute them across a swarm. These sub-agents can asynchronously explore vast solution spaces, cross-verify information, and debate conflicting strategies.
\end{itemize}

Ultimately, scaling sleep-time compute redefines the lifecycle of LLM agents. Agents become persistent, self-improving collaborators that continuously learn, adapt, and validate across both user and environment dimensions. Moreover, this approach aligns with cognitive models of human expertise: just as individuals consolidate memories and refine mental models during rest~\cite{klinzing2019mechanisms}, agents can leverage downtime to refine policies and simulate future trajectories. In this sense, sleep-time scaling serves as the computational analog of cognitive rehearsal and long-term memory consolidation.


\subsubsection{Human Time: Scaling Multi-Step Capabilities}
A key consideration in evaluating the deployment potential of LLM agents is the concept of baseline human time, denoted as $T_0$. This represents the time required for a human to complete a given task. Understanding the characteristics of $T_0$ is crucial for identifying domains where LLM agents can provide meaningful time savings. The following properties of $T_0$ are particularly relevant:

\begin{itemize}[leftmargin=*]
\item \textit{Context-Dependence:} The value of $T_0$ is inherently sensitive to the context in which a task is performed. A task that is trivial in a static environment may become significantly more time-consuming when subjected to environmental constraints or divided attention. For instance, as discussed in Section~\ref{sect_usability_trend}, ordering a coffee via a mobile application may take mere seconds when a user is seated comfortably at home. However, performing the same action while driving drastically inflates $T_0$ due to safety constraints and limited interaction bandwidth.

\item \textit{Personalization and Accessibility:} The baseline human time $T_0$ varies widely across individuals. Expertise, prior knowledge, tool familiarity, and cognitive style all influence task duration. What is trivial for an expert may be slow and error-prone for a novice. This variability is particularly critical for elderly users or individuals with disabilities, who may require extended time to navigate standard digital interfaces due to motor or visual impairments. In this light, LLM agents offer a pathway to digital equity~\cite{resta2008issues}.

\item \textit{Non-Linearity of Task Composition:} As tasks become more compound---involving multiple steps, interdependencies, and decision points---$T_0$ tends to rise super-linearly. This non-linearity arises from the cognitive switching penalty~\cite{schmitz2023task}: the mental effort required to maintain working memory of the overall goal while executing individual sub-steps. For humans, multi-step tasks necessitate sustained planning, state tracking, and context switching, all of which increase the probability of fatigue-induced errors.
\end{itemize}

Reflecting on the observations in Section~\ref{sect_usability_trend}, it is clear that LLM agents are most impactful in domains where the baseline human time $T_0$ is relatively high. These domains typically involve contextually demanding and multi-step tasks. To operate effectively in these high-$T_0$ scenarios, LLM agents must scale their multi-step capabilities. Key requirements include:

\begin{itemize}[leftmargin=*]
\item \textbf{Long-Horizon Reasoning:} Maintaining coherence, consistency, and goal alignment across extended interactions and multi-turn reasoning sequences.
\item \textbf{State Persistence:} Accurately retaining and transferring relevant information across steps without hallucination, drift, or data loss. This is essential for preserving task continuity and avoiding redundant sub-tasks.
\item \textbf{Self-Correction:} As the probability of failure compounds in longer reasoning chains, agents must be able to detect, localize, and correct intermediate errors to avoid cascading failures and improve reliability.
\end{itemize}

The ability to handle such multi-step, stateful tasks will be a defining feature of next-generation LLM agents and a central driver of their economic value. Currently, however, agent capabilities degrade drastically as the number of steps increases~\cite{lu2024weblinx}.


\subsubsection{Agent Interaction Time: Scaling Proactive Interaction}
Agent interaction time, denoted as $T_{Agent}$, represents the total time a human user spends engaging with an agent to complete a task. Unlike agent latency, which refers to computational processing time, $T_{Agent}$ is a measure of human effort. This includes all phases of interaction: initial prompt formulation, intent clarification, iterative refinements (\eg, rephrasing, follow-up questions), and final result verification. 

To reduce agent interaction time, systems must shift the burden of context acquisition from the user to the agent~\cite{lu2024proactive}. This requires a transition from reactive execution to proactive interaction, where the agent actively infers goals, anticipates needs, and automatically completes the task. This shift is tightly coupled with scaling sleep-time compute, allowing the agent to reason and prepare before user interactions occur. Reducing $T_{Agent}$ can be achieved through three key mechanisms:

\begin{itemize}[leftmargin=*]
\item \textbf{Intent Inference and Goal Prediction:} Agents should be capable of inferring user intent from partial, implicit, or ambiguous inputs. Rather than relying on the user to fully articulate the task, agents can leverage interaction history and environmental cues to predict likely goals. Clarifying questions should be asked only when necessary, minimizing interaction turns. However, as shown in \citet{lu2024proactive}, current LLM agents still fall short in predictive capabilities and often fail to anticipate user goals effectively, highlighting a key area for improvement.

\item \textbf{Implicit Context Resolution:}
In traditional reactive paradigms, users are responsible for explicitly verbalizing the state of the world. Proactive agents can mitigate this by integrating multimodal inputs (\eg, voice, imagery, and structured UI elements) to automatically capture environmental and task-specific context~\cite{durante2024agent}. Moreover, agents can reduce input overhead by offering intelligent autocompletions, prefilled templates, and context-aware suggestions. These features lower interaction costs, especially in repetitive or structured tasks, and enable faster transitions from intent to execution.

\item \textbf{Proactive Verification and Trust:} 
The time allocated to result verification is inversely proportional to user trust. When trust is low, users scrutinize every detail and mentally simulate the agent's steps, effectively duplicating the computational effort. To compress this verification phase, agents must proactively perform self-verification before presenting the final output. This includes running internal consistency checks and cross-referencing sources. By exposing a concise, human-readable reasoning trace, the agent transforms verification from a forensic audit into a lightweight sanity check.
\end{itemize}

The long-term goal of scaling proactive interaction is to drive $T_{Agent}$ toward zero for routine and well-understood tasks. In such cases, the user's role shifts from active instruction to strategic oversight and high-level delegation.


\subsection{Scaling Down for Cost Reduction} 
While recent advances stem from scaling up for information gain, we argue that this expansion must be followed by a phase of scaling down to optimize unit utility. Once a capability threshold is met, the focus shifts from maximizing performance at any cost to minimizing the resources required to maintain it. We discuss critical strategies for lowering the operational burden of LLM agents without compromising output quality.

\begin{itemize}[leftmargin=*]
    \item \textbf{From Real-time Thinking to Memory Retrieval:} Agents relying solely on real-time inference incur high computational penalties, especially when regenerating reasoning chains for similar tasks~\cite{wu2025unlocking}. By integrating long-term memory systems, agents can bypass redundant computation through retrieval-based reasoning, transforming time complexity from expensive generation to efficient lookup. This mirrors human cognitive efficiency~\cite{gabrieli1998cognitive, evans2003two}: experts solve tasks rapidly not because they think faster, but because they remember better. Shifting from compute-bound inference to memory retrieval dramatically reduces latency and cost, particularly in repetitive or domain-specific applications.
    
    \item \textbf{From Generalist Giants to Specialized Experts:} While large generalist models define the state-of-the-art, they are often over-parameterized for routine sub-tasks. Scaling down involves compressing these capabilities into compact, specialized architectures through knowledge distillation~\cite{xu2024survey}. Techniques such as quantization, pruning, and speculative decoding allow these models to operate within tighter resource budgets, significantly lowering the barrier to deployment~\cite{egashira2024exploiting}.
    
    \item \textbf{Hardware-Software Co-optimization:} Infrastructure dictates the lower bound of operational cost. Advances in AI-specific hardware (\eg, Groq~\cite{gwennap2020groq}, Cerebras~\cite{lie2022cerebras}) and inference-optimized software stacks (\eg, vLLM~\cite{kwon2023efficient}, FlashAttention~\cite{dao2022flashattention}) directly alleviate bottlenecks. Improving agent efficiency at scale requires the co-evolution of software and hardware, moving beyond raw FLOPs to metrics that matter for interactive agents, such as cost per task.
\end{itemize}

\section{Call to Action}
To close the usability gap and unlock the next wave of adoption, we must reframe the development and deployment of LLM agents around maximizing Agentic ROI.

\textbf{First, we call for a redesign of agents intended for the mass market, everyday applications.} Today’s LLM agents thrive in high-ROI domains such as coding and research, where expert users tolerate back-and-forth interactions in exchange for significant utility gains. However, to serve the broader public, agents must become more robust, proactive, and accessible. This means designing systems that reduce interaction friction and anticipate user intent. By shifting the focus from raw model capability to real-world usability, we can create agents that are not only powerful, but also intuitive and dependable.

\textbf{Second, to make informed and scalable deployment decisions, we must adopt Agentic ROI as a first-class evaluation metric.} Unlike traditional benchmarks, Agentic ROI captures the tangible value delivered to users by balancing improvements in information quality and time savings against the cost of agent usage. Its utility spans the ecosystem: for researchers, it offers a benchmark that aligns model progress with real-world human benefit; for developers, it guides optimization across quality, latency, and computational efficiency; for organizations, it provides a framework for assessing the economic viability and practical impact of agent integration; and for users, it ensures that systems are responsive to their time, expectations, and constraints.

We encourage the community to collaborate on shared datasets, benchmarks, and tools for estimating Agentic ROI across tasks, domains, and user populations. This shift from abstract performance scores to metrics grounded in lived experience is essential for meaningful progress.

LLM agents are at a pivotal moment. The foundational capabilities are already in place: we now have general-purpose models with unprecedented capacity for reasoning, planning, and acting. But realizing their full potential depends not on scaling up indiscriminately, but on scaling wisely to prioritize information gain, time savings, and cost. Now is the time to close the usability gap---by building agents that are not only intelligent, but truly useful.

\section{Alternative Views}\label{sect:alternative}
While we advocate for a framework grounded in Agentic ROI, we must address alternative perspectives shaping current deployment strategies. Below, we re-examine two common views through this lens.

\begin{itemize}[leftmargin=*]
    \item \textbf{View 1: Repetitive tasks automatically warrant LLM agent deployment.}
This view rests on the intuition that automation is most valuable when tasks are repetitive, and thus, LLM agents should naturally be applied in such contexts.

\textit{Response:} Repetition only increases ROI if it yields time savings and quality at a reasonable cost. Many repetitive tasks are better served by traditional rule-based automation or lightweight models. LLM agents here may be overengineered and expensive. Worse, if the agent introduces variability or errors in an otherwise deterministic process, the resulting verification costs can negate any time savings.

\item \textbf{View 2: Model capability and accuracy determine usability.} 
This view assumes that higher raw model performance automatically translates to a more usable agent.

\textit{Response:} Capability is merely one component of usability. A powerful model requiring complex prompting or heavy verification offers low practical usability. Conversely, a modest but responsive and intuitive agent may deliver higher net utility. Moreover, because accuracy gains often incur higher computational costs, Agentic ROI diminishes when marginal improvements fail to offset the expense, thereby reducing practical usability.

\end{itemize}

\section{Conclusion}\label{sect:conclusion}
In this position paper, we argue that the key barrier to the practical usability of LLM agents lies not in model capability alone, but in the agent's ability to deliver utility relative to the costs incurred during real-world use. To formalize this tradeoff, we introduce and operationalize the concept of Agentic ROI. Agentic ROI quantifies the ratio of information gain and time savings to the cost. Applying this metric, we reveal a usability gap, noting that domains with the highest market demand show significantly lower Agentic ROI. We analyze development trends in optimizing Agentic ROI and outline future directions for bridging the usability gap, paving the way for LLM agents that are truly usable, scalable, and accessible in real-world applications.

\bibliography{ref}
\bibliographystyle{icml2026}

\newpage
\appendix
\onecolumn
\section{Literature Review}
\subsection{Agent Capabilities}
Recent advancements in LLMs have catalyzed a shift from static chatbots to autonomous agents capable of multi-step reasoning and tool execution. Early foundational work, such as the ReAct framework \cite{yao2022react}, demonstrated that interleaving reasoning traces with action execution significantly improves performance. This paradigm has been further expanded by multi-agent architectures like AutoGen \cite{wu2023autogen} and MetaGPT \cite{hong2023metagpt}, which utilize role-playing and collaborative dialogue to solve complex software engineering and data analysis problems.

The capability ceiling of these agents has risen dramatically. Benchmarks such as GAIA \cite{mialon2023gaia} and $\tau^2$-Bench \cite{barres2025tau} evaluate agents on tasks requiring deep reasoning and real-world tool usage, with recent models achieving success rates that were previously unattainable. Furthermore, the integration of memory modules \cite{park2023generative} and retrieval-augmented generation~\cite{lewis2020retrieval} has allowed agents to maintain long-term context, theoretically enabling them to handle continuous, open-ended tasks. However, these improvements in agent capabilities do not necessarily guarantee real-world usability.

\subsection{Agent Applications} 
LLM agents have been applied to various domains ranging from software development~\cite{novikov2025alphaevolve} and scientific research~\cite{huang2025deep} to specialized fields such as education~\cite{chu2025llm} and healthcare~\cite{yang2025medaide}.

One of the most mature applications of LLM agents is in software development. Agents are increasingly deployed to handle end-to-end coding tasks, moving beyond simple code completion to complex repository-level reasoning. Frameworks such as SWE-agent~\cite{yang2024swe} and CodeAgent~\cite{zhang2024codeagent} demonstrate the ability to autonomously navigate file systems, debug errors, and execute shell commands to resolve GitHub issues. 

In the scientific domain, agents act as copilots for research, automating the tedious aspects of literature review and data synthesis. Systems like ChemCrow~\cite{bran2023chemcrow} integrate LLMs with domain-specific tools to plan chemical syntheses and operate robotic lab equipment. Similarly, AI co-scientist~\cite{gottweis2025towards} shows potential in drug repurposing, novel target discovery, and explaining mechanisms of bacterial evolution and anti-microbial resistance. 

Moreover, agents are reshaping general enterprise workflows. By integrating with APIs and SaaS platforms, agents can orchestrate complex business processes, including supply chain monitoring, customer support, and automated presentation generation~\cite{zheng2025pptagent}. Unlike traditional Robotic Process Automation (RPA)~\cite{hofmann2020robotic}, which relies on rigid scripts, LLM-based agents can adapt to unstructured inputs and edge cases, offering a more resilient solution for dynamic business environments~\cite{pruucha2025llm}.
Despite these broad applications, the transition from controlled benchmarks to complex, real-world environments remains a significant challenge.

\section{Broader Impact}
The broader impact of Agentic ROI can be understood across several dimensions.

\textbf{Guiding Product Design.} The Agentic ROI framework serves as more than a descriptive metric for the usability gap; it provides a prescriptive guide for practitioners. By quantifying the trade-offs between cost and utility, it helps designers optimize agent interfaces, workflows, and interaction paradigms for maximum user value.
    
\textbf{Understanding Human-AI Co-Evolution.} From a longer-term perspective, the human baseline time $T_0$ may not be static; rather, it co-evolves with agent interaction time $T_{Agent}$ as part of a broader human-AI ecosystem. For example, users may gradually offload cognitive tasks to agents, reducing their own proficiency, thereby increasing the human baseline time. 
    
\textbf{Democratizing AI Access.} Focusing on Agentic ROI highlights usability barriers for non-expert users. By revealing where the costs outweigh the time savings, it guides efforts toward greater inclusivity, making LLM agents more accessible to the general public, not just technical experts. 
    
\textbf{Alternative Research Directions.} This perspective invites a shift in research priorities---from pure model performance to end-to-end utility. It encourages interdisciplinary approaches that integrate economics and systems thinking. Furthermore, it opens avenues for exploring multi-agent coordination, where Agentic ROI must account for the overhead of inter-agent communication and synchronization.


\end{document}